\documentclass[conference]{IEEEtran}
\IEEEoverridecommandlockouts
\usepackage{cite}
\usepackage{amsmath,amssymb,amsfonts}
\usepackage{graphicx}
\usepackage{textcomp}
\usepackage{algorithm}
\usepackage{algpseudocode}
\usepackage{mathrsfs}
\usepackage{subfigure}
\usepackage{adjustbox}

\usepackage{xcolor}
\def\BibTeX{{\rm B\kern-.05em{\sc i\kern-.025em b}\kern-.08em
    T\kern-.1667em\lower.7ex\hbox{E}\kern-.125emX}}
\begin{document}

\title{Communication Trade-offs in Federated Learning of Spiking Neural Networks}

\author{\IEEEauthorblockN{ Soumi Chaki}
\IEEEauthorblockA{\textit{National Supercomputer Centre} \\
\textit{Linköping University}\\
Linköping, Sweden \\
soumi.chaki@liu.se}
\and
\IEEEauthorblockN{David Weinberg}
\IEEEauthorblockA{\textit{Hitachi Energy} \\
Ludvika, Sweden \\
david.o.weinberg@hitachienergy.com}
\and
\IEEEauthorblockN{Ayca Özcelikkale}
\IEEEauthorblockA{\textit{Department of Electrical Engineering} \\
\textit{Uppsala University}\\
Uppsala, Sweden \\
ayca.ozcelikkale@angstrom.uu.se}
}

\maketitle

\begin{abstract}
Spiking Neural Networks (SNNs) are biologically inspired alternatives to conventional Artificial Neural Networks (ANNs). Despite promising preliminary results, the trade-offs in the training of SNNs in a distributed scheme are not well understood.  Here, we consider SNNs in a federated learning  setting where a high-quality global model is created by aggregating multiple local models from the clients without sharing any data.  
We investigate federated learning for training multiple SNNs at clients when two mechanisms reduce the uplink communication cost: i) random masking of the model updates sent from the clients to the server; and ii) client dropouts where some clients do not send their updates to the server. We evaluated the performance of the SNNs using a subset of the Spiking Heidelberg digits (SHD) dataset.
The results show that a trade-off between the random masking and the client drop probabilities is crucial to obtain a satisfactory performance for a fixed number of clients.
\end{abstract}


\section{Introduction}
Spiking neural networks (SNNs) have gained considerable interest among researchers for their ability to solve real-world problems in the deep learning paradigm while improving their energy efficiency and latency \cite{Lee2016}. 
SNNs have obtained comparable performance to standard artificial neural networks (ANNs) or conventional convolutional neural networks (CNNs) on different benchmarked datasets such as MNIST and NMNIST \cite{Lee2016}. Hence, SNNs show great potential for the implementation of on-device low-power online learning and inference \cite{Skatchkovsky2020}. 

With the availability of large amounts of data and advanced computation devices, Federated Learning (FL) has gained interest among researchers \cite{Li2020, McMahan2017}. In a federated learning  setting,  a high-quality global model is created by aggregating multiple local models from the clients without sharing any data.
 SNNs implemented in an FL setup are promising in terms of  exploiting SNN's energy efficiency while preserving data security as the dataset associated with a particular node is not shared with the server or other nodes, only model parameters are exchanged. 
Accordingly, in \cite{Skatchkovsky2020}, an online FL-based learning rule i.e. \textit{FL-SNN} is introduced to train several online SNNs simultaneously. Some recent papers show the performances of federated SNNs to solve real-world datasets \cite{Venkatesha2021,liu2022,xie2022}, including CIFAR10, CIFAR100, and BelgiumTS image datasets. In particular, in \cite{Venkatesha2021}, a $15\%$ increase in overall accuracy was reported with $4.3$ times energy efficiency with CIFAR10 and CIFAR100 datasets while using  SNNs instead of ANNs in a federated setup.

Although  the costs for uplink and downlink communications between the server and the nodes can be expensive in an FL setup and hence been the focus of various work in the context of ANNs, see for instance \cite{BecirovixChenLarsson_2022, HuChenLarsson_2021, Chen_2021, Yang_2021} or adaptive filters \cite{Danaee_2022}, this issue has not been investigated in the context of SNNs. 
Another important issue is the unresponsive nodes. Although initial communication trade-offs in terms of unresponsive nodes have been investigated \cite{Venkatesha2021} in the context of SNNs,  at the moment it is not clear to which extent the communication cost can be reduced, either by masking or dropping updates  and to which extend the unresponsive nodes can be tolerated without compromising the performance of the federated SNNs significantly.
In this paper, we address this gap.

We consider the training of SNNs in a federated setting under uplink communication costs. We investigate how the communication cost can be reduced by restricting the updates to be a sparse matrix during uplink communication (masking). We explore the performance trade-offs between the number of clients and the amount of masking of the model updates during uplink communication. We also investigate dropout where some clients do not send their updates to the server. We investigate to which extent the unresponsive nodes, i.e. dropouts, can be tolerated while working with a fixed number of clients.
We evaluated the performance of the SNNs using a subset of the Spiking Heidelberg digits (SHD) dataset \cite{cramer2020heidelberg}.
The results show that although some masking and dropout can be tolerated by the SNNs, a trade-off between the masking and the client drop probabilities is crucial to obtain a satisfactory performance for a fixed number of clients.

\section{SNNs and Federated Learning}
\subsection{Spiking Neural Network}
\begin{figure}[htbp]
\centerline{\includegraphics[width=0.8\columnwidth]{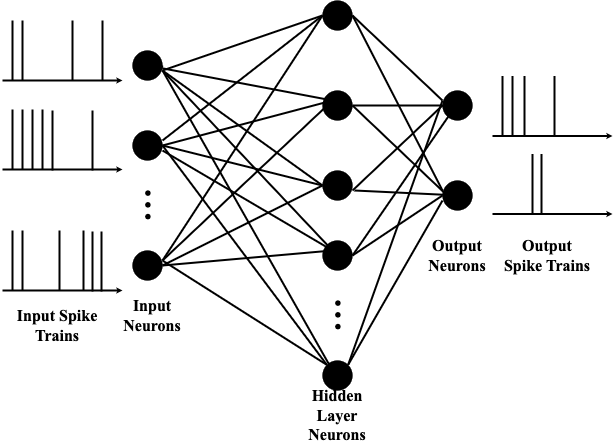}}
\caption{Schematic diagram of a Spiking Neural Network with one hidden layer.}
\label{fig_snn_architecture}
\end{figure}

Fig.~\ref{fig_snn_architecture} shows a schematic diagram of an SNN with a single hidden layer. The network has multiple input and hidden layer neurons and two output layer neurons in Fig.~\ref{fig_snn_architecture}. 
Each neuron accumulates the incoming spikes and scales by their corresponding synaptic weights and generates its membrane potential. When its membrane potential reaches a certain threshold, a spike is generated. This mechanism is called Integrated-and-Fire (IF). For the Leaky-Integrate-and-Fire (LIF) variant, after generating a spike the membrane potential also leaks at a constant rate \cite{Venkatesha2021, Neftci2019}. Following a spike, membrane potential is reduced to a resting potential or reduced by the threshold value. 
Let us consider, an infinitely short current pulse as
\begin{equation}
I(t)=q\delta(t)
\end{equation}
when a spike starts at t=0, where $q$ is the amount of charge in the current pulse and $\int_{- \infty}^{\infty}\delta(t)=1$. 
The dynamics of the synaptic current $I_{i}^{syn}(t)$ for neuron $i$ in the hidden layer, where $S_{j}^{IN}(t)$ is the spike train from input neuron $j$, $\omega_{i,j}^{h}$ is the weight, and $\tau_{syn}>0$ is the time constant of the exponential decay of the synoptic current, can be presented as:
\begin{equation} \label{eq:i_cont}
\tau_{syn}\frac{dI_{i}^{syn}(t) }{dt}=-I_{i}^{syn}(t)+\sum_{j}^{}\omega_{i,j}^{h}qS_{j}^{IN}(t).   
\end{equation}
The dynamics of the membrane potential $V_{i}^{}(t)$ of hidden neuron $i$ where  $I_{i}^{syn}(t)$ can be presented by:
\begin{equation} \label{eq:v_cont}
\tau_{mem}\frac{dV_{i}^{syn}(t) }{dt}=-(V_{i}^{}(t)-V_{rest}^{})+RI_{i}^{syn}(t)   
\end{equation}
where, $R$ is the membrane resistance, $V_{rest}$ is the resting potential.
In this article, we use a discrete-time formulation of SNNs.
Using a small time-step $\Delta_{t} >0$, and setting $\frac{q}{\tau_{syn}}=1$, (\ref{eq:i_cont}) can be approximated as  \cite{Neftci2019} 
\begin{equation} \label{eq:i_discrete}
I_{i}^{syn}[m+1]=\alpha I_{i}^{syn}[m]+\sum_{j}^{}\omega_{i,j}^{h}S_{j}^{IN}[m],    
\end{equation} 
where $m$ is the time-index, $S_{j}^{IN}[m] \in {0,1}$, and decay parameter of the current $\alpha = exp(-\frac{\Delta_{t}}{\tau_{syn}})(0<\alpha<1)$.
Setting $R=1$ and $V_{rest}=0$, the membrane potential in discrete form can be written as \cite{Neftci2019}
\begin{equation} \label{eq:v_discrete}
V_{i}[m+1]=\beta V_{i}[m]+ I_{i}^{syn}[m]   
\end{equation}
where, $\beta=exp(-\frac{\Delta_{t}}{\tau_{mem}})$ is the decay parameter for the voltage. Similarly, the membrane potential for the output layer can be calculated using (\ref{eq:i_discrete})-(\ref{eq:v_discrete}) by carrying out appropriate modifications.
\\ The discontinuous nature of the spiking non-linearity poses a challenge in training SNNs. Instead of using the actual derivative of the spike, it is replaced by a surrogate function and the choice of the function is not unique as the resulting surrogate gradient \cite{Zenke2021}. Surrogate gradient methods help overcome the difficulties associated with discontinuous non-linearity \cite{Neftci2019}. 

\begin{figure}[htbp]
\centerline{\includegraphics[width=\columnwidth]{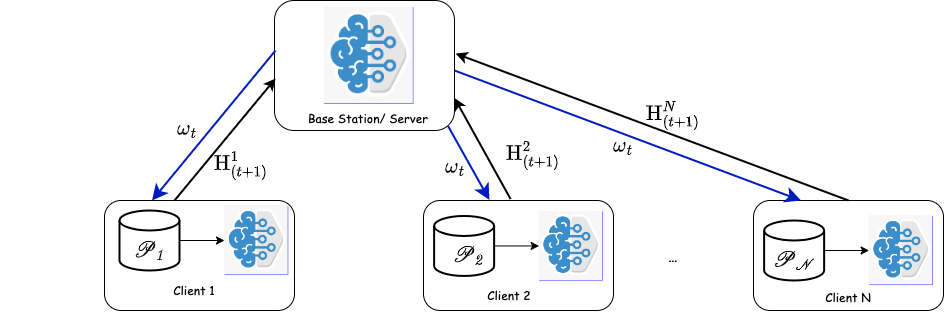}}
\caption{Schematic diagram of Federated Learning.}
\label{fig_federated_learning}
\end{figure}

\subsection{Federated Learning}
A typical FL system consists of a server or base station and $N$ number of clients. At the beginning of the training, the server broadcasts an initial global model $\omega_{0}$ to the $N$ clients. The initial model can be a model with heuristically selected parameters or a model trained on a public dataset. Each client $k = 1,2,..., N$ is then locally trained a model using the dataset available at a particular client and the model update is communicated periodically to the server. Each iteration of sending the model updates from clients to the server is called a round. Therefore, only model updates are shared instead of the client dataset retaining the privacy of the clients. Then, the server aggregates the model updates and creates an updated global model for broadcasting to the server for the next round. 

\section{FL-SNN-MaskedUpdate Scheme}
\subsection{Updates in Federated Learning Scheme}
At round $t$, the broadcasted model $\mathrm{\omega}_{t}^{k}$ from the server to all the clients is the same for all clients $k$ i.e. $\mathrm{\omega}_{t}$. At each node $k$, an updated model $\mathrm{\omega}_{t+1}^{k}$ is created after training with the local datasets and $\mathrm{H}_{t+1}^{k}$ is the difference between the locally updated model $\mathrm{\omega}_{t+1}^{k}$ and the broadcasted model $\mathrm{\omega}_{t}^{k}$
\begin{equation}
\mathrm{H}_{(t+1)}^{k}=\mathrm{\omega}_{(t+1)}^{k}-\mathrm{\omega}_{t}^{k}
\end{equation}
At round $t$, the model update $\mathrm{H}_{(t+1)}^{k}$ from a particular client $k$ is sent to the server. \textit{FedAvg} \cite{McMahan2017}, \textit{FedProx} \cite{sahu2018convergence}, and \textit{FedMA} \cite{wang2020federated} are some of the algorithms used for aggregating the local model updates $\mathrm{H}_{t+1}^{k}$ at the server to create the global model $\mathrm{\omega}_{t+1}$ after round $t$. In case of \textit{FedAvg}, the server aggregates the model updates based on a weighted average of the updates depending on the number of data samples at a particular client with respect to the total number of data samples combining all clients. We use $FedAvg$ in this paper. Fig.~\ref{fig_federated_learning} illustrates the process of federated learning.

\subsubsection{Random Masking}
The federated learning process involves major communication costs when sending model updates from clients to the server (uplink communication) and  broadcasting a global model from the server to the clients (downlink communication). In \cite{konevcny2016federated}, a structured update process was introduced by using a random mask to reduce the uplink communication costs in an FL setup. In the case of a random mask, a model update is restricted to be a sparse matrix, adhering to a pre-determined random sparsity pattern (using a random seed). The pattern is independently generated for each client in each round. Let us assume, $m$ (in $\%$) is the amount of masking to the model update $\mathrm{H}_{(t+1)}^{k}$ by a particular client $k$ at the end of round $t$. So, instead of sending the whole model update from a client to the server, only the non-zero entities of the update $\mathrm{H}_{(t+1)}^{k}$ along with the seed $s_{(t+1)}^{k}$ can be sent to the server after round $t$. The server can reconstruct a sparse model update $\mathrm{\tilde{H}}_{(t+1)}^{k}$ for each client $k$ at a particular round $(t+1)$ using the seed $s_{(t+1)}^{k}$ and non-zero entries $\left[\mathrm{H}_{(t+1)}^{k}\right]_{\text{nz}}$. Then, the server creates the global model for the next round ($t+1$) i.e. $\omega_{(t+1)}$ by aggregating $\mathrm{\tilde{H}}_{(t+1)}^{k}$
 as follows: 
\begin{equation}
H_{(t+1)}\equiv \frac{1}{N_c}\sum_{k=1}^{N_c}\mathrm{\tilde{H}}_{(t+1)}^{k}
\end{equation}
where $N_c$ is the number of clients.

\subsubsection{Client Dropout}
In case of malfunction of some of the nodes, we assume $N_{c}$ number of clients out of total $N$ clients are working and sending $\left[\mathrm{H}_{(t+1)}^{k}\right]_{\text{nz}}$ and seeds $s_{t}^{k}$ to the server and $N_{c} \leq N$. Then, the server will create $\omega_{(t+1)}$ and broadcast it to the clients using the available $\left[\mathrm{H}_{(t+1)}^{k}\right]_{\text{nz}}$  and seeds.

\subsection{FL-SNN}
The complete pseudo-code of our developed scheme is presented in Algorithm~\ref{alg:flsnnmaskedupdate}. Let us assume, $\mathscr{P}_{k}$ is the dataset available at client $k$. For each client, the dataset $\mathscr{P}_{k}$ 
is divided into several batches of size $B$ which are used to train the machine learning model for each local epoch $e$. For our case, the number of local epochs $E$ is set as $1$, therefore, each data sample passes through the ML model only once. The working $N_c$ clients are indexed by $k$ and $\eta$ is the learning rate at the clients.  

\algrenewcommand\algorithmicfunction{}
\begin{algorithm}
\caption{\textit{FL\textendash SNN\textendash MaskedUpdate}}\label{alg:flsnnmaskedupdate}
  \begin{algorithmic}[1]
    \Function{\textbf{ServerExecutes}}{} \Comment{Run on the server}
    \State initialize  $\omega_{0}$
    \For {each round $t=1,2,...,d$}
        \For {each working client $k=1,2,...,N_c$}
        \State $\left[\mathrm{H}_{t+1}^{k}\right]_{\text{nz}}$, $s_{t+1}^{k}$ = ClientUpdateMasked({$\mathrm{\omega}_{t}^{k},m$})
        \State construct $\mathrm{\tilde{H}}_{(t+1)}^{k}$ using $\left[\mathrm{H}_{(t+1)}^{k}\right]_{\text{nz}}$, $s_{(t+1)}^{k}$
        \EndFor
        \State $H_{(t+1)}\equiv \frac{1}{N_c}\sum_{k=1}^{N_c}\mathrm{\tilde{H}}_{(t+1)}^{k}$
        \State $\omega_{(t+1)}=\omega_{t}+H_{(t+1)}$
    \EndFor    
    \State \textbf{broadcast} $\omega_{(t+1)}$ to the $N_c$ clients
    \EndFunction
    \\
    \Function{\textbf{ClientUpdateMasked}}{$\mathrm{\omega}_{t}^{k}$,$m$} \Comment{Run on client $k$}
    \For {epoch $e=1,2,...,E$} %{\texttt{<$1\le e \le E$>}}
        \For {\texttt{$b \in \beta$}}
        \State $\mathrm{\omega}_{(t+1)}^{k} \gets \mathrm{\omega}_{t}^{k} - \eta \nabla l(\mathrm{\omega}_{t}^{k};b)$
        \EndFor 
    \EndFor   
    \State $\mathrm{H}_{(t+1)}^{k}=\mathrm{\omega}_{(t+1)}^{k}-\mathrm{\omega}_{t}^{k}$
    \State generate a seed number $s_{t+1}^{k}$ for client $k$, round $t+1$
    \State for $s_{(t+1)}^{k}$, generate $\left[\mathrm{H}_{(t+1)}^{k}\right]_{\text{nz}}$ and its indices for $m$.
    \State \textbf{return} $\left[\mathrm{H}_{(t+1)}^{k}\right]_{\text{nz}}$, $s_{(t+1)}^{k}$ to the server.
    \EndFunction
  \end{algorithmic}
\end{algorithm}

\section{Numerical Results}
\subsection{Dataset}
%\subsubsection{Dataset} 
We have implemented the developed algorithm on a subset of SHD dataset \cite{cramer2020heidelberg}. In the original SHD dataset, the total number of training and testing samples are around $10k$ combining the samples from all $20$ classes together. The dataset consists of high-quality studio recordings of spoken digits from $0$ to $9$ in both German and English languages.  
The detailed descriptions of the data collection, processing, and analysis can be found in \cite{cramer2020heidelberg}. 
In the experiments, we have used the samples pertaining to the first five labels (labels $0-4$) of the SHD dataset.
The number of training and testing samples for our experiments were $2011$ and $534$ respectively.

\subsection{Training Procedure}
The experiments were carried out using python software installed on a Linux server Dell PowerEdge R$740$ with intel Xeon Gold $6248$R @ $3$Ghz ($48$ cores) CPU, $2$ x NVIDIA Quadro RTX $6000$ $24$GB GPU and $768$GB RAM. 
We have maintained the structure of the SNNs as a single hidden layer SNN and kept the parameter values the same throughout the experiments. We have used back-propagation with surrogate gradient \cite{Neftci2019} and ADAM algorithm to train the SNNs \cite{kingma2015adam}. The number of rounds for the federated learning setup is $150$. Table~\ref{tab1:SNNparameters} lists the parameters associated with SNNs and they are chosen heuristically. It took around $1.2$ hour time to complete training testing for a particular experiment, e.g. for a fixed number of nodes (say $4$ nodes) for a particular random masking ($10\%$) level for $150$ communication rounds.
\begin{table}[htbp]
\caption{Parameter values of the Spiking Neural Networks for all experiments}
\label{tab1:SNNparameters}
\begin{center}
\begin{tabular}{|c|c|}
\hline
\textbf{Parameter}                                                                                          & \textbf{Values} \\ \hline
Number of input nodes                                                                                       & 700             \\ \hline
Number of time-samples                                                                                      & 100             \\ \hline
Number of hidden nodes                                                                                      & 50              \\ \hline
Number of output nodes                                                                                      & 5               \\ \hline
Batch size (Train/Test)                                                                                     & 20              \\ 

\hline
Number of local epochs                                                                                      & 1               \\ \hline
Learning rate                                                                                               & 0.0001          \\ \hline
Decay parameter $\alpha$ for current                                                                                 & 0               \\ \hline
Decay parameter $\beta$ for voltage                                                                                 & 1               \\ \hline
\begin{tabular}[c]{@{}c@{}}Weight Initialization  \\ (mean/scale factor in standard deviation)\end{tabular} & 0/1             \\ \hline
\end{tabular}
\label{tab1}
\end{center}
\end{table}

We investigated the performances of the high-quality global models generated at the server for the different numbers of clients varying between $2$ to $10$. The amount of masking during the uplink communication also varied from no masking to a small amount of masking ($10\%$) and finally a very high amount of masking ($98\%$). 

\subsection{Learning Curves} 
\begin{figure}[htpb]
    \centering
    \subfigure[]
    {\includegraphics[width=0.75\columnwidth,trim=8mm 3mm 1mm 8mm, clip=true]{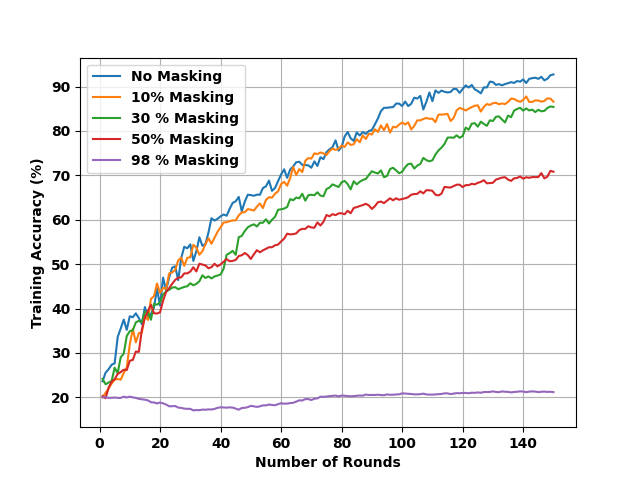}
        \label{fig:nodevsmask_training}}
    % \\
    \vspace{-3mm}
    \subfigure[]
    {\includegraphics[width=0.75\columnwidth,trim=8mm 3mm 1mm 8mm, clip=true]{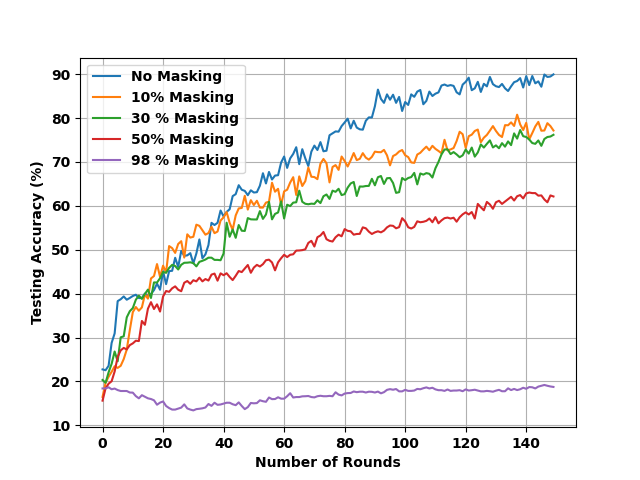}
        \label{fig:nodevsmask_testing}}
    
    \caption{The effect of masking of model updates on model performances over $150$ rounds with $4$ clients}
    \label{fig:nodevsmask}
\end{figure}

The performances of the federated SNNs are shown in Fig.~\ref{fig:nodevsmask} in terms of (a) training and (b) testing accuracy values for $4$ clients and $150$ rounds.
The global models continue to improve over the rounds when working without any masking or with a small amount of masking to the model updates during training and testing both. On the other hand, almost no improvement over the performances is visible for a very high amount of masking ($98\%$). The model performances change significantly when the amount of masking varies from a very low amount (e.g. $10\%$) to a high amount (e.g. $50\%$) in contrast to the insignificant performance reduction from no masking to only $10\%$ masking.

\subsection{Trade-offs between the number of clients and the amount of random masking }
\begin{figure}[htpb]
    \centering
    \subfigure[]
    {
        \includegraphics[width=0.75\columnwidth,trim=3mm 0.5mm 0.5mm 6mm, clip=true]{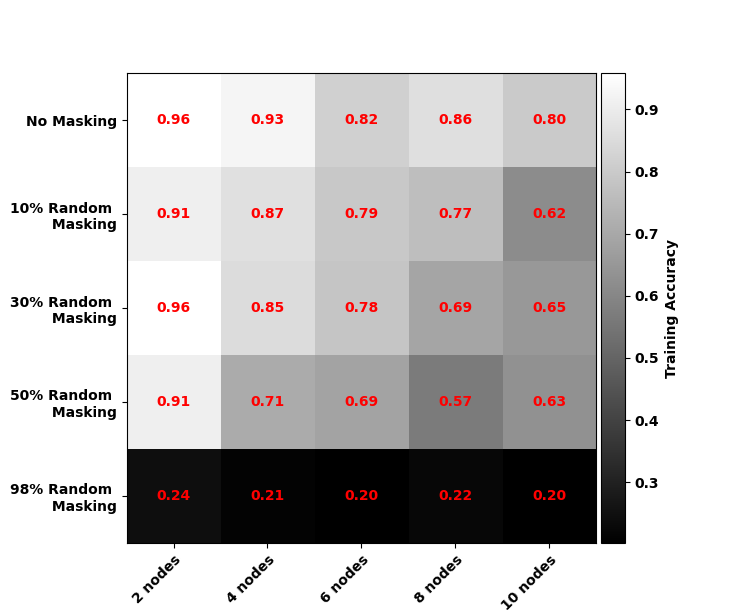}
        \label{fig:heatmap1_training}
    }
    \\
    \subfigure[]
    {
        \includegraphics[width=0.75\columnwidth,trim=3mm 0.5mm 0.5mm 6mm, clip=true]{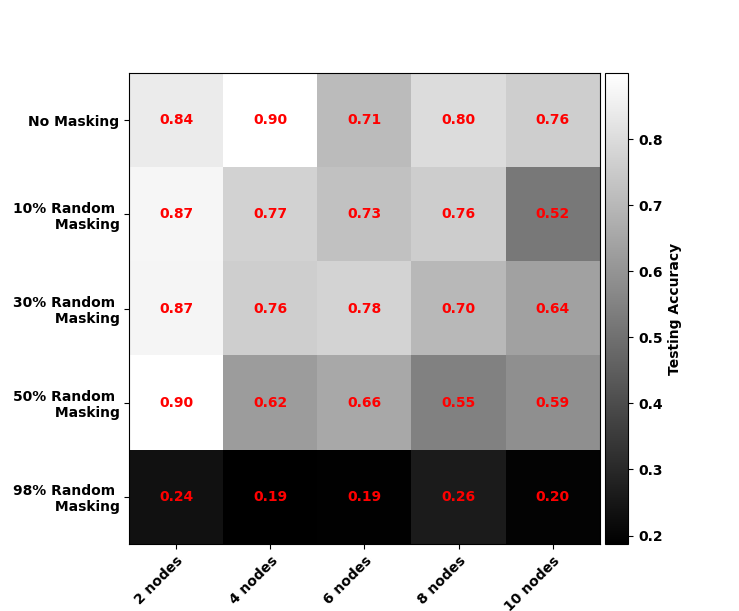}
        \label{fig:heatmap1_testing}
    }
    
    \caption{The effect of masking with respect to the number of clients for 150 rounds}
    \label{fig:nodevsmaskHeatmap}
\end{figure}

The combined effect of the number of clients and the amount of masking on training and testing performances of the global models after $150$ communication rounds are presented using the heatmaps in Fig.~\ref{fig:nodevsmaskHeatmap}. The accuracy values vary between $\sim 0.2$ to near $1$ and are presented using the gray colormaps with the final values  mentioned in red. After each round, the updated global model is saved. The training and testing performances of the global models over the rounds are obtained by using the $150$ saved global models on the complete training and test datasets. The global models for the setup with only two clients performed better for different masking levels compared to the setups with a higher number of clients. It can be seen from Fig.~\ref{fig:nodevsmaskHeatmap} that sometimes the performance improved with the increase of masking (e.g. Fig.~\ref{fig:nodevsmaskHeatmap}(a)-(b), for $2,8$ and $10$ nodes). It shows the regularization effect on the SNNs exploiting its spike-dependent and sparsity-driven characteristics similar to biological systems. Some of the recent papers \cite{han2022adaptive,yan2022backpropagation} are developing training algorithms to introduce regularization to already sparse, low-energy consuming SNNs. Our results would be beneficial for providing guidelines for those algorithms in terms of applicable sparsity level. 
\subsection{Effect of malfunctioning nodes}
\begin{figure}[htbp]
\centerline{\includegraphics[width=0.75\columnwidth,trim=3mm 0.5mm 0.5mm 6mm, clip=true]{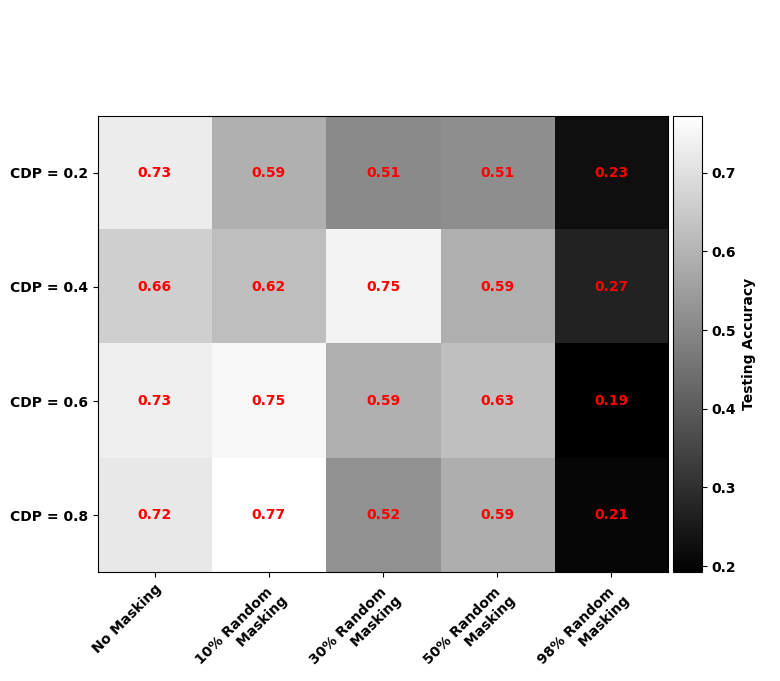}}
\caption{Testing accuracy values of global models for different level of masking and client drop probabilities}
\label{fig:heatmap2testacc}
\end{figure}

We explored the effect of malfunctioning nodes, i.e. drop-out,  in addition to the amount of masking for $10$ clients over $150$ rounds. The performances of the saved global models on the testing dataset are demonstrated in Fig.~\ref{fig:heatmap2testacc}. The client drop probability (CDP) is varied between $0.2$ to $0.8$, where $0.2$ means that $2$ out of a total of $10$ clients stopped working at each round. The testing accuracy is $\sim 0.2$ i.e. random for $98\%$ masking irrespective of the CDP values, which is expected. The dropping of nodes along with random masking of the model updates is introducing regularization effects to the system, resulting in interesting observations. For example, the accuracy with CDP$=0.4$ and $30\%$ masking obtained better performance that a smaller amount of masking with the same CDP values. The performances with a smaller number of nodes (with higher CDP values) are relatively better. 

\section{Conclusions and Future Work}
We investigated trade-offs for reducing uplink communication in a federated SNN setup using random masking of the model updates and implemented it on a benchmarked spiking dataset \textemdash SHD (Labels $0$ to $4$). In general, the performance of global models for a fixed number of nodes and rounds deteriorates with an increasing  amount of random masking. However, our results also show that masking may improve the overall performance instead of deteriorating it. Due to the spike-based and sparsity-driven nature of SNNs, masking can be interpreted as inducing regularization of the system. Important future research directions include downlink communication constraints as well as other communication channel imperfections. 

\bibliographystyle{IEEEtran}
\bibliography{Project_SNN}

\end{document}